# Mobile Robot Navigation Using Hand-Drawn Maps: A Vision Language Model Approach

Aaron Hao Tan, *IEEE Student Member*, Angus Fung, *IEEE Student Member*, Haitong Wang, *IEEE Student Member*, and Goldie Nejat, *IEEE Member*

*Abstract*— Hand-drawn maps can be used to convey navigation instructions between humans and robots in a natural and efficient manner. However, these maps can often contain inaccuracies such as scale distortions and missing landmarks which present challenges for mobile robot navigation. This paper introduces a novel Hand-drawn Map Navigation (HAM-Nav) architecture that leverages pre-trained vision language models (VLMs) for robot navigation across diverse environments, hand-drawing styles, and robot embodiments, even in the presence of map inaccuracies. HAM-Nav integrates a unique Selective Visual Association Prompting approach for topological map-based position estimation and navigation planning as well as a Predictive Navigation Plan Parser to infer missing landmarks. Extensive experiments were conducted in photorealistic simulated environments, using both wheeled and legged robots, demonstrating the effectiveness of HAM-Nav in terms of navigation success rates and Success weighted by Path Length. Furthermore, a user study in real-world environments highlighted the practical utility of hand-drawn maps for robot navigation as well as successful navigation outcomes compared against a non-hand-drawn map approach.

## I. INTRODUCTION

Mobile robot navigation tasks may have to be performed in environments that can change, for example, due to structural instability in search and rescue scenarios [1], construction progress during renovations or a new build [2], and retail store reconfiguration [3]. In order for robots to navigate these environments they either use map-based [4], [5] or map-less [6], [7] methods. In map-based methods, accurate maps are generated prior to navigation using human teleoperation [8] or autonomous robot exploration [9]. However, map acquisition can be: 1) costly and time consuming [10], and 2) requires expert knowledge [11]. On the other hand, map-less methods [6], [7] can represent a robot's environment using real-time sensory data. However, existing map-less methods

This work was supported in part by the Natural Sciences and Engineering Research Council of Canada (NSERC), and in part by the Canada Research Chairs program (CRC).The authors are with the Autonomous Systems and Biomechatronics Laboratory (ASBLab), Department of Mechanical and Industrial Engineering, University of Toronto, Toronto, ON M5S 3G8, Canada (e-mail: aaronhao.tan@utoronto.ca;angus.fung@mail.utoronto.ca*;* haitong.wang@mail.utoronto.ca;nejat@mie.utoronto.ca). Corresponding author: Aaron Hao Tan.

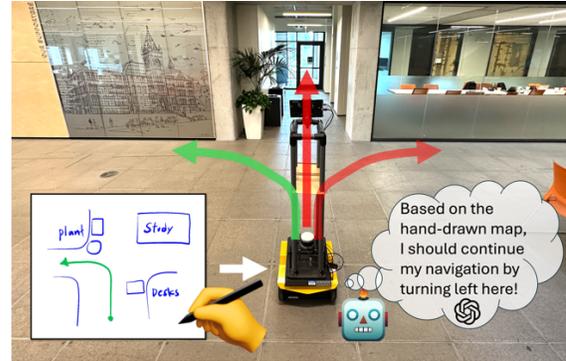

**Fig. 1.** An overview of mobile robot navigation with a hand-drawn map.

require simultaneous exploration and navigation to reach the robot's goal position. This can affect navigation efficiency in terms of the total robot distance traveled due to the initial exploration stage [6].

Robot navigation using hand-drawn maps can provide an alternative approach to both map-based and map-less navigation methods, Fig. 1. Hand-drawn maps are freehand sketches generated by people based on their memory of an environmental layout to represent spatial relationships within the robot's environment [12]. Thus, hand-drawn maps can be used effectively for robot navigation without the need for a priori resource-intensive map acquisition [13], [14] or simultaneous exploration during navigation [15].

To date, existing robot navigation methods using hand-drawn maps can be classified as: 1) heuristic methods [13], [16], [17], [18], [19], [20], [21], which recognize known landmarks for execution of predefined actions; and 2) probabilistic methods [9], [10], [15], [22], [23], [24], [25], [26], which match sensory data with map features for localization. However, these methods have been restricted to simple environments with hand-crafted landmarks, (e.g., boxes and cylinders), which do not represent complex real-world environments with realistic landmarks (e.g., furniture) [24] and multi-level floors. They have also required accurate hand-drawn maps with precise spatial layout representation [9], which are hard to obtain due to variations in human memory when sketching [24].

In this paper, we present a novel hand-drawn map robot navigation architecture, HAM-Nav, which uses pre-trained vision language models (VLMs) to interpret visual and textual cues from hand-drawn maps for robot navigation in unknown environments. HAM-Nav is the first method to generalize across diverse environments and varying hand-drawing styles without task-specific training. Our main research contributions are: 1) the introduction of a new adaptive visual prompting method, Selective Visual Association Prompting (*SVAP*), that places the robot view alongside a dynamically updated topological map overlaid on top of the hand-drawn map. Using *SVAP*, HAM-Nav can estimate the robot's position and select appropriate navigation actions in a zero-shot manner by enabling pre-trained VLMs to directly associate environmental features with corresponding elements in the hand-drawn map; and 2) the development of a Predictive Navigation Plan Parser (*PNPP*) to infer missing landmark information (e.g., class and location) using the common-sense knowledge of pre-trained VLMs to account for human errors in a hand-drawn map.

## II. RELATED WORKS

We categorized robot navigation using hand-drawn maps into: 1) heuristic methods [13], [16], [17], [18], [19], [20], [21], and 2) probabilistic methods [9], [10], [15], [22], [23], [24], [25], [26]. We also review visual language navigation methods [27], [28], [29], [30], [31], [32], [33].

### A. Heuristic Methods

Heuristic methods use predefined patterns, such as spatial proximity, to interpret geometric features in hand-drawn maps and convert them into navigation commands. These include rule-based [21], fuzzy control [13], [16], [17], [18], and optimization-based [19], [20] approaches.

Rule-based methods compute objectness scores by combining image segmentation with fuzzy c-means clustering to extract features from hand-drawn maps, which are then used to guide robot navigation using sensory feedback [21].

Fuzzy control methods convert pixel coordinates into spatial descriptions, classifying them as polygons (landmarks) or line segments (paths) [13], [16], [17], [18]. Landmark positions are inferred using polygon boundary forces and mapped to actions through fuzzy inference rules.

Optimization methods use quadratic programming to compute waypoints by modeling spatial relationships between landmarks and paths as virtual springs. Minimizing potential energy here yields navigation trajectories [19], [20].

### B. Probabilistic Methods

Probabilistic methods apply statistical models to interpret hand-drawn maps and localize robots. Approaches include Hidden Markov Models (HMM) [22], Monte Carlo Localization (MCL) [10], [15], [23], [24], [25], Bayesian Filtering (BF) [9], and Supervised Learning (SL) [26].

In [22], a variable-duration HMM was trained to recognize strokes and inter-strokes in hand-drawn maps, generating robot motion vectors. MCL methods use particle filtering to localize a robot by: 1) aligning the hand-drawn map to the environment layout [15], or 2) estimating map deformations [10], [23], [24], [25].

In [9], a BF method generated a local occupancy grid from panoramic RGB images. The robot's belief state was updated by comparing this grid to a hand-drawn map using similarity scores, and navigation vectors were computed via grid search.

The SL method in [26] used a CNN to predict control points for aligning the robot's view with the hand-drawn map via alpine-based registration. The A* algorithm was then used to plan a path to the goal.

### C. Visual Language Navigation Methods

Visual Language Navigation (VLN) guides robots in unseen environments using natural language and visual inputs. Recent methods leverage VLMs for generalization across diverse instructions and scenes. Approaches include: 1) building topological maps from language and visual data to encode spatial relations [27], 2) parsing instructions into actions using navigation history and object detections [28], [29], [30], and 3) integrating models for parsing, perception, and trajectory summarization for collaborative decision-making [31], [32], [33].

### D. Summary of Limitations

Heuristic methods use reactive control to execute fixed actions and cannot adapt to environmental changes, relying only on a limited set of primitive landmarks. Probabilistic methods require hand-drawn maps that accurately reflect geometry and scale, where deviations in these maps lead to localization errors. All methods are limited to single-floor settings as they assume a 2D hand-drawn map can directly align with sensor data. However, in multi-floor environments, representing stacked floors on a 2D plane distorts spatial relationships. These methods also assume all landmark shapes and positions are correctly drawn, which is unrealistic due to human error [24]. VLN methods require users to verbally provide complete navigation instructions in a single attempt. However, recalling accurate instructions in complex environments during a single utterance is difficult, as it requires users to mentally reconstruct the spatial configuration of the environment without visual feedback (e.g., hand-drawn map).

To address the above limitations, we propose HAM-Nav, an architecture that uniquely leverages VLMs to: 1) detect realistic landmarks and enable zero-shot navigation in multi-floor environments by using adaptive visual prompting to align visual features from the environment with textual and spatial cues in the hand-drawn map, and 2) account for human errors in hand-drawn maps by using co-occurrence landmark patterns to predict missing landmarks.

## III. ROBOT NAVIGATION WITH HAND-DRAWN MAPS PROBLEM DEFINITION

The robot navigation problem using hand-drawn maps consists of requiring a mobile robot to autonomously navigate from a given starting position to a desired position within an unknown environment, utilizing navigation instructions conveyed through a free hand sketch. This sketch, referred to as a hand-drawn map, $\mathcal{M}_h$, is generated by a person based on their memory. $\mathcal{M}_h = (\mathcal{S}_h, \mathcal{L}_h, \mathcal{P}_h)$ consists of three components, Fig. 2(a): 1) the spatial configuration, $\mathcal{S}_h$, which represents the outer boundary and structural layout of the robot's environment; 2) the landmarks, $\mathcal{L}_h = (\mathcal{L}_h^c, \mathcal{L}_h^\ell)$, which consist of text descriptions that depict a landmark's class (e.g., chair, desk), $\mathcal{L}_h^c$, and its pixel location within $\mathcal{M}_h$, $\mathcal{L}_h^\ell$; and 3) a path, $\mathcal{P}_h$. The path includes the initial robot position, $p_0 = (x_0, y_0)$, Node 1 in Fig. 2(a), and the desired robot position $p_d = (x_d, y_d)$, Node 4 in Fig. 2(a). Note, possible errors in landmark positions, distances and scaling can exist because of a person's imperfect recollection of the environment. A person may also misplace or omit landmarks in $\mathcal{M}_h$, leading to an incomplete set for $\mathcal{L}_h$ compared to ground truth.

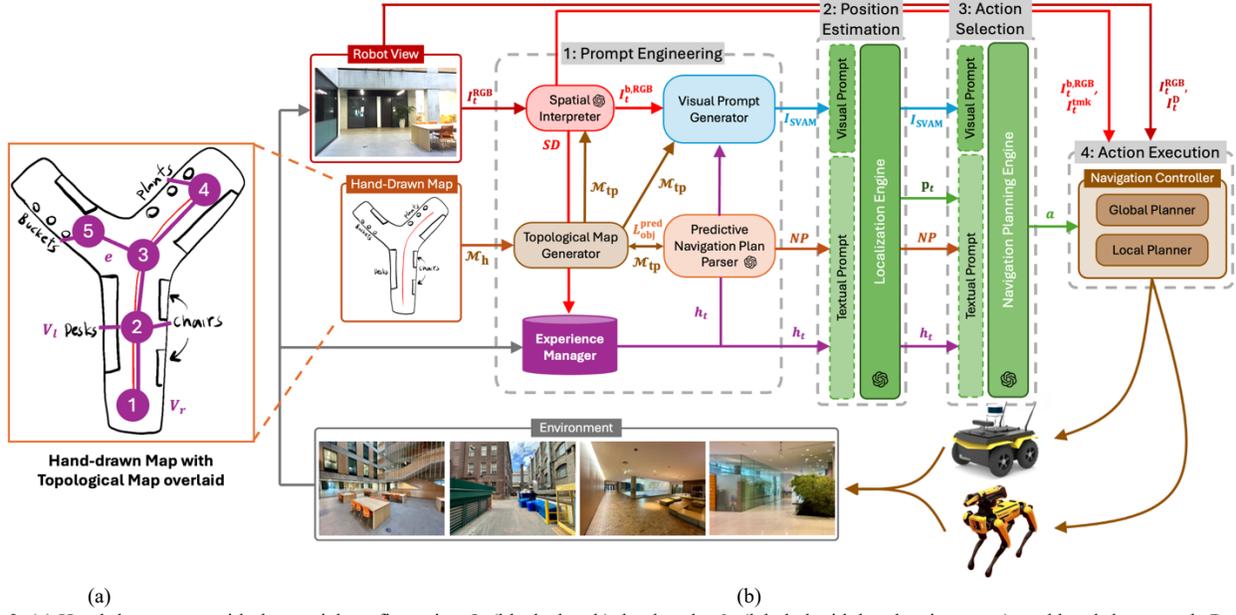

(a)          (b)

**Fig. 2.** (a) Hand-drawn map with the spatial configuration $\mathcal{S}_h$ (black sketch), landmarks $\mathcal{L}_h$ (labeled with hand-written text), and hand-drawn path $\mathcal{P}_h$ (red line) overlaid with the topological map $\mathcal{M}_{tp}$ (purple line); and (b) the proposed HAM-Nav architecture. ⓖ denotes a VLM.

The mobile robot has an onboard RGB-D camera, $C(t)$, to capture both RGB, $I_t^{RGB}$, and depth images, $I_t^D$ of its surroundings. The objective is to solve the following robot navigation problem: Given $\mathcal{M}_h$ and $\mathcal{P}_h$, a mobile robot must localize itself within $\mathcal{M}_h$ and generate a sequence of actions $a(t)$ in order to navigate to $p_d$ based on real-time observations $(I_t^{RGB}, I_t^D)$ from $C(t)$. The sequence of actions $a(t)$ is determined by a robot action function $f(\cdot)$, Eq. 1. The goal is for the robot to reach $p_d$ at the final timestep, $T$.

$$a(t) = f(\mathcal{M}_h, \mathcal{P}_h, C(t)) \text{ s.t. } p(T) = p_d. \quad (1)$$

## IV. HAND-DRAWN MAP NAVIGATION ARCHITECTURE

The proposed Hand-Drawn Map Navigation Architecture (HAM-Nav) is presented in Fig. 2(b) and consists of the following four stages. **Stage 1: Prompt Engineering** consists of a *Topological Map Generator* (*TMG*), *Spatial Interpreter* (*SI*), *Visual Prompt Generator* (*VPG*), *Predictive Navigation Plan Parser* (*PNPP*), and an *Experience Manager* (*EM*) to extract navigation and environmental features from $\mathcal{M}_h$ and $I_t^{RGB}$ in order to structure and generate visual and textual prompts. **Stage 2: Position Estimation** uses a *Localization Engine* (*LE*) to estimate the robot's position $p_t$ within $\mathcal{M}_h$ based on the visual and textual prompts generated in Stage 1. **Stage 3: Action Selection** consists of a *Navigation Planning Engine* (*NPE*) to select an embodiment-agnostic discrete navigation action $a$ based on $p_t$. Lastly, **Stage 4: Action Execution** uses a *Navigation Controller* (*NC*) to convert $a$ into robot velocities to be executed in the environment. The following details the modules within each stage. Table I provides a summary of important notations used throughout the proposed methodology

TABLE I: Summary of Important Notations

| Notation | Description |
|---|---|
| $\mathcal{M}_h, \mathcal{M}_{tp}$ | The hand-drawn map provided by the user, and the topological map generated based on the hand-drawn map |
| $\mathcal{S}_h, \mathcal{L}_h, \mathcal{P}_h$ | Hand-drawn spatial structure, landmarks, and path within $\mathcal{M}_h$ |
| $V_r, V_l$ | Robot position nodes and landmark nodes |
| $L_{obj}, L_{obj}^{pred}$ | Object landmarks and predicted object landmarks |
| $h_t \in H$ | Past navigation experiences |
| $SD, p_t, a$ | Scene description, estimated robot position and action. |
| $I_t^{RGB}, I_t^D, I_t^{b,RGB}$ | RGB image, depth image, and a labeled RGB image with bounding boxes for detected objects |
| $I_t^{tmk}$ | Traversable region mask image |
| $I_{SVAM}, NP$ | Selective visual association prompt and navigation plan |
| $\sigma_{vis}, \sigma_{text}$ | Visual prompt, and textual prompt functions for VLM |

### A. Topological Map Generator (TMG)

The *TMG* creates a topological map, $\mathcal{M}_{tp}$, based on $\mathcal{M}_h$ for robot localization and navigation planning. Namely, $\mathcal{M}_{tp} = (V, E)$ is a graph where $V = V_r \cup V_l$ represents the set of vertices comprising robot position nodes $V_r$, landmark nodes $V_l$, and a set of edges connecting these nodes, $E$, Fig. 2(a). $\mathcal{M}_{tp}$ is provided to the following modules: 1) *SI* to perform landmark detection, 2) *VPG* for visual prompt generation, and 3) *PNPP* to predict missing landmarks for path planning. The predicted landmarks $L_{obj}^{pred}$ are passed back to *TMG* to update the $\mathcal{M}_{tp}$ via the bidirectional connection shown in Fig. 2(b). This update enables $\mathcal{M}_{tp}$ to include both landmarks from $\mathcal{M}_h$, using optical character recognition, and $L_{obj}^{pred}$.

### B. Spatial Interpreter (SI)

At each timestep $t$, *SI* generates: 1) a labeled image, $I_t^{b,RGB}$, that includes the bounding boxes and object classes for the detected landmarks, and 2) a textual description, $SD$, of $I_t^{RGB}$. Two types of landmarks are detected within $I_t^{RGB}$: 1) object landmarks, $L_{obj}$, which include physical objects such as furniture and vehicles, and 2) structural landmarks, $L_{str}$, such as multi-way junctions (e.g., left and right turns). $L_{obj}$ are detected with Grounding DINO [34], using landmark classes from $V_l$ in $\mathcal{M}_{tp}$. $L_{str}$ are detected using our own three-stage approach, Fig. 3. Firstly, Grounded-Segment Anything Model [35] is used to segment the traversable region within $I_t^{RGB}$ by generating a pixel-level mask, $I_t^{tmk}$. In the second stage, edges are extracted from $I_t^{tmk}$ to generate $I_t^e$ using the Hough Transform. Edges in $I_t^e$ are grouped into four different categories based on their orientation and length: horizontal $I_t^{e,h}$, vertical $I_t^{e,v}$, positive slope $I_t^{e,p}$, or negative slope $I_t^{e,n}$. Lastly, left and right turns are identified by using a rule-based approach that checks the intersection patterns between $I_t^{e,h}$, $I_t^{e,v}$, $I_t^{e,p}$, $I_t^{e,n}$. Intersections that match spatial configurations of turns based on visual perspective are classified as left or right. In Fig. 3, the bounding boxes for both $L_{obj}$ (pink) and $L_{str}$ (green) are shown in $I_t^{b,RGB}$. The bounding box coordinates is denoted by $L_{cd}$.

The textual description of $I_t^{RGB}$, $SD$, at each $t$ is obtained using a VLM. This process involves both a visual, $\sigma_{vis}(\cdot)$, and a textual, $\sigma_{text}(\cdot)$, prompt. Due to page limitation, all system prompts used in this work are provided in our YouTube video (link in Section V). Specifically, $\sigma_{text}(\cdot)$ taks as input $L_{dict}$, which describes the generalized landmark locations relative to the robot's perspective (e.g., left, front, right) within $I_t^{RGB}$. To obtain generalized landmark locations, $I_t^{RGB}$ is divided into three equal-width horizontal quadrants: left $Q_l$, front $Q_f$, and right $Q_r$. Each detected landmark is assigned to a quadrant based on the x-coordinate of its bounding box center, $L_{cd}^i$. These assignments are stored in a dictionary $L_{dict} = \{L_{obj}: Q\}$. This dictionary is used to generate a structured textual prompt in the format "<$L_{obj}$> on your <$Q$>", which is provided to the VLM to produce a detailed description $SD$ of $I_t^{RGB}$. This process is described by:

$$SD = \text{VLM}\left(\sigma_{vis}(I_t^{RGB}), \sigma_{text}(L_{dict})\right). \quad (2)$$

The $I_t^{b,RGB}$ is used by the *VPG* module for visual prompt generation and *NC* module for robot navigation, while $SD$ is used by the *EM* module to retrieve relevant experiences.

### C. Experience Manager (EM)

The *EM* module collects and retrieves past navigation experiences to provide historical contextual navigation information using Retrieval Augmented Generation (RAG). The entire set of historical experiences, denoted as $H$, is stored onboard the robot, with each specific experience $h_t$, representing the navigation data at a particular $t$. Each $h_t$ includes the prior observation $SD'_t$, estimated robot position $p'_t$, and executed action $a'_t$, at the corresponding $t$. To retrieve the most relevant $h_t$ from $H$, the cosine similarity is computed between the current observation and all stored experiences. The $h_t$ with the highest similarity is used as a textual prompt to the *LE* module for robot position estimation and the *NPE* module for navigation planning.

### D. Visual Prompt Generator (VPG)

We developed the *VPG* module to enable Selective Visual Association Prompting by generating a visual prompt, $I_{SVAM}$, that determines the relationship between the features in $I_t^{RGB}$ and $\mathcal{M}_h$, Fig. 4. Specifically, $I_{SVAM}$ consists of an RGB image with two side-by-side components: 1) $I_t^{b,RGB}$, and 2) a pruned topological map, $\mathcal{M}'_{tp}$, overlaid on top of $\mathcal{M}_h$. Herein, $\mathcal{M}'_{tp}$ contains only the robot position node candidates with the highest likelihoods of representing the robot's true position in the environment. These likelihoods are determined using a probabilistic model that prunes position node candidates with low retention probability $\zeta(v_i)$. The $\zeta(v_{r,i})$ for each $v_{r,i} \in V_r$ is determined by the following logistic function:

$$\zeta(v_{r,i}) = \frac{1}{1 + e^{\alpha \cdot (d(v_{r,i}, p') - \beta) + \gamma \cdot \delta(v_{r,i}, p', a')}}, \quad (3)$$

where $\alpha$ is a weighting factor that influences $\zeta(v_{r,i})$ based on $d(v_{r,i}, p')$, while $\beta$ is a sensitivity parameter $\zeta(v_{r,i})$ with respect to distance. The transition function $\delta(v_{r,i}, p', a')$ determines the probability of the robot arriving at $v_i$ after executing action $a'$ at $p'$. $\gamma$ is a weighting factor influencing $\zeta(v_{r,i})$ based on $\delta$. A logistic function was chosen for its probabilistic interpretation, tunable parameters, and computational efficiency [36]. The topological map $\mathcal{M}'_{tp}$ includes only nodes where $\zeta(v_{r,i})$ exceeds 0.5. We set $\beta = 2$ and both $\alpha$ and $\gamma$ to 0.5 based on expert-guided tuning to prune nodes unlikely to represent the robot's position. $I_{SVAM}$ is used by the *LE* and *NPE* modules for robot position estimation, and action selection, respectively.

### E. Predictive Navigation Plan Parser (PNPP)

We uniquely propose a *PNPP* to provide: 1) predicted landmarks, $L_{obj}^{pred}$, and 2), a textual description of the navigation plan, $NP$. Specifically, the *PNPP* infers omitted landmarks using the VLM. The input to the VLM includes both a visual $\sigma_{\text{vis}}(\mathcal{M}_h)$ and a textual $\sigma_{\text{text}}(V)$ prompt, which are conditioned on $\mathcal{M}_h$, and all $V$ in $\mathcal{M}_{\text{tp}}$, respectively. The VLM uses this information to infer potential co-occurring landmarks based on the spatial relationships and proximity of nearby landmarks. The landmark prediction process for a node $v_i \in V$ is as follows:

$$L_{\text{obj}}^{\text{pred}} = \text{VLM}(\sigma_{\text{vis}}(\mathcal{M}_h), \sigma_{\text{text}}(V)). \quad (4)$$

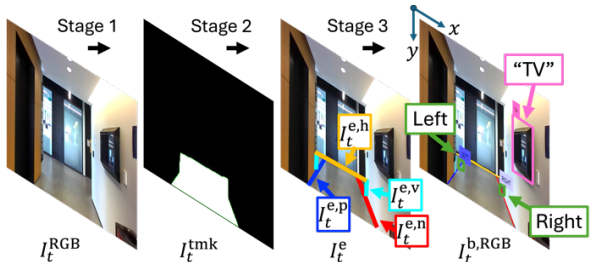

**Fig. 3.** The three stages of structural landmark detection.

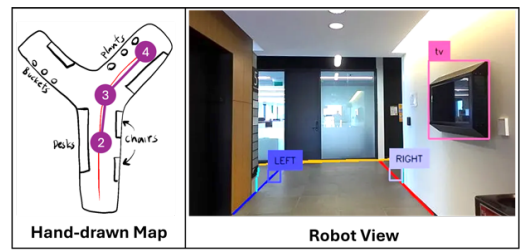

**Fig. 4.** A visual prompt, $I_{SVAM}$, that consists of a hand-drawn map with a pruned topological map, $\mathcal{M}'_{tp}$, and the robot view, $I_t^{b,RGB}$.

The $L_{\text{obj}}^{\text{pred}}$ for each $v_i$ is incorporated into $\mathcal{M}_{\text{tp}}$ by the *TMG* to be used by the *SI* module for landmark detection.

To generate $NP$, $\mathcal{M}_{\text{tp}}$ is segmented into local segments $S_i$ by junction nodes (nodes with a change in direction), $V_{\text{junc}} \subseteq V$, in the environment, shown in Fig. 5(a). For each $S_i$, the associated $L_{\text{str}}, L_{\text{obj}}, L_{\text{obj}}^{\text{pred}}$ are obtained from $\mathcal{M}_{\text{tp}}$. A descriptive sentence is generated for each $S_i$ following a fixed structure: "<u>*navigation action*</u> pass the <u>*landmarks*</u>, and <u>*navigation action*</u> when you see <u>*landmarks*</u>", as illustrated in Fig. 5(b). The collection of all local navigation plans $\{NP_i\}$ forms the global navigation plan $NP$, which is provided as textual prompt to *LE* and *NPE* modules.

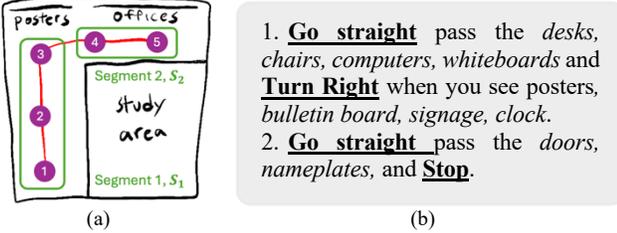

(a)  (b)

**Fig. 5.** (a) $\mathcal{M}_{\text{tp}}$ overlaid on top of $\mathcal{M}_h$, with robot position nodes (purple) and landmark nodes (handwriting). The green boxes represent segments $S_i$; and (b) the local navigation plan, $NP_i$. Predicted landmarks $L_{\text{obj}}^{\text{pred}}$ are italicized.

### F. Localization Engine (LE)

The *LE* module uses the VLM to estimate the robot's current position, $p_t$, by selecting a robot position node $v_i$ in $\mathcal{M}'_{\text{tp}}$. The input to the VLM includes both visual, $\sigma_{\text{vis}}(I_{\text{SVAM}})$, and textual prompts, $\sigma_{\text{text}}(SD', p', a', NP)$. We utilize two prompting techniques for the textual prompt to estimate $p_t$. The first is chain of thought prompting (CoT) [37], to decompose the robot position estimation task into smaller explicit steps. This is achieved by asking the VLM to perform step-by-step reasoning by first identify visual landmarks and then relating these visual landmarks to the hand-drawn map before generating $p_t$. Score-based prompting (SB) [38] is used by asking the VLM to explicitly generate the probabilities of the robot position estimations. The position estimation process is:

$$p_t = \text{VLM}(\sigma_{\text{vis}}(I_{\text{SVAM}}), \sigma_{\text{text}}(SD', p', a', NP)_{\text{CoT,SB}}). \quad (5)$$

The estimated $p_t$ is used by the *NPE* module for robot navigation action selection.

### G. Navigation Planning Engine (NPE)

The objective of the *NPE* is to generate embodiment-agnostic high-level actions such as "move forward", "turn right", "turn left", and "stop" using the VLM. The *NPE* module, like the *LE*, uses $\sigma_{\text{vis}}(I_{\text{SVAM}})$ and $\sigma_{\text{text}}(SD', p', a', NP, p_t)$ for zero-shot navigation decision making. CoT [37] and SB [38] prompting was used to guide the reasoning process. Specifically, CoT decomposes the navigation task into first understanding $p_t$, in $\mathcal{M}_h$, and then relating $p_t$ to $NP$. CoT prompting is used to make this reasoning process explicit and sequential, while SB prompting is used to assign a probability score to each possible action, representing the likelihood of an action to successfully complete the navigation plan. The action with the highest probability score, $a$, is then selected and passed to the *NC* module for execution. The action selection process is:

$$a = \text{VLM}(\sigma_{\text{vis}}(I_{\text{SVAM}},), \sigma_{\text{text}}(SD', p', a', NP, p_t)_{\text{CoT,SB}}). \quad (6)$$

### H. Navigation Controller (NC)

The *NC* module converts $a$ into robot velocities $(v, \omega)$ for navigation execution in three stages using $I_t^{\text{RGB}}$, $I_t^{D}$, $L_{\text{cd}}$ and $I_t^{\text{tmk}}$. First, the centroids pixel coordinates of $I_t^{\text{tmk}}$ are selected for the "move forward" action. For the "turn left" and "turn right" actions, the detected $L_{\text{str}}$ coordinates from $L_{\text{cd}}$ are selected. Second, the selected 2D pixel coordinates are projected into a 3D goal coordinates, to be passed to the robot path planners, using the pinhole camera model [39]. Third, a global path planner generates waypoints towards the 3D goal coordinates, and a local planner converts these waypoints in to $(v, \omega)$ for robot navigation.

## V. EXPERIMENTS

We conducted two sets of experiments to evaluate the performance of our novel HAM-Nav architecture: 1) an ablation study to assess the contributions of the specific design choices of HAM-Nav, and 2) a user study to investigate the feasibility and usability of HAM-Nav approach in real-world environments.

### A. Ablation Study in Simulated Environments

Navigation performance was investigated using four metrics: 1) navigation time (NT) in seconds to reach the desired position $p_d$, 2) navigation distance (D) in meters to reach $p_d$, 3) success weighted by path length (SPL), to evaluate the robot's navigation path compared to the human hand-drawn path, and 4) success rate (SR).

*1) Simulated Environments:* Two 3D photorealistic environments were generated in the Gazebo simulator, Fig. 6.

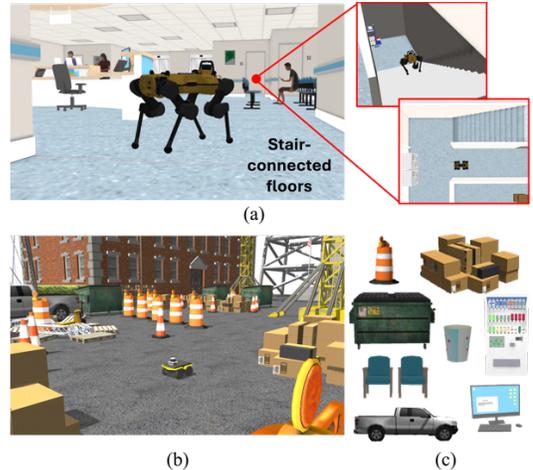

**Fig. 6.** Simulated environment of: (a) an indoor multi-floor workplace ($25\ m \times 55\ m$), (b) an outdoor construction site ($40\ m \times 40\ m$), and (c) examples of photorealistic landmarks for both environments.

TABLE II: Ablation Study with Averaged Results

| Methods | NT (s) ↓ | D (m) ↓ | SR ↑ | SPL ↑ |
|---|---|---|---|---|
| HAM-Nav (ours) | 634 | 28 | 80% | 0.712 |
| HAM-Nav w/o $L_{\text{dict}}$ | 743 | 34 | 45% | 0.327 |
| HAM-Nav w/o $L_{\text{obj}}^{\text{pred}}$ | 780 | 36 | 40% | 0.287 |
| HAM-Nav w/o $\mathcal{M}'_{\text{tp}}$ | 893 | 41 | 25% | 0.134 |
| HAM-Nav w/o $EM$ | 1583 | 72 | 5% | 0.013 |
| HAM-Nav w Q2.5-3B | 879 | 40 | 28% | 0.18 |
| HAM-Nav w Q2.5-7B | 794 | 35 | 38% | 0.27 |
| HAM-Nav w Q2.5-72B | 707 | 30 | 65% | 0.58 |

The first environment was a structured indoor multi-floor workplace featuring rectilinear walls and stair-connected floors, Fig. 6(a). The second environment was an unstructured outdoor construction site with irregular navigation paths formed by randomly placed landmarks, Fig. 6(b). Examples of landmarks in these environments included pylons, boxes, dumpsters, chairs and computers, Fig. 6(c).

*2) Mobile Robots:* A Clearpath Jackal wheeled robot with a differential drive system, and a Boston Dynamic Spot quadruped robot were deployed. Both robots have an onboard RGB-D sensor. For the Jackal robot, the A* algorithm global planner and the Timed Elastic Band (TEB) local planner [40] were used. For the Spot robot, the Rapidly-exploring Random Trees based global planner (RRT) [41], and a non-linear model predictive controller (NMPC) local planner [42] were used. At the time of writing, we used GPT-4o for its multimodal reasoning, low latency, and reliable API [43]. However, future implementations are encouraged to incorporate newer models as they become available.

*3) Ablation Study Methods:* We compared against: (1) HAM-Nav without $L_{\text{dict}}$ to assess the impact of generalized landmark locations in $SD$, (2) HAM-Nav without $L_{\text{obj}}^{\text{pred}}$ to evaluate the effect of the predicted landmarks, (3) HAM-Nav without $\mathcal{M}'_{\text{tp}}$ to investigate the contribution of pruned topological maps, (4) HAM-Nav without $EM$ to determine the significance of historical navigation information, and (5) HAM-Nav with Qwen 2.5-VL models (Q2.5) [44] to evaluate performance with varying sizes of open-source VLMs.

*4) Procedure:* For each environment, two hand-drawn maps were created per robot platform with two random start and goal positions. Five trials were conducted per map. For SPL calculation, $\mathcal{P}_h$ was converted to an equivalent path in the metric map as the optimal reference. A trial was successful if the robot reached the goal within 0.5 meters.

*5) Results:* Table II presents the averaged NT, D, SR, and SPL for HAM-Nav and its ablations. The full HAM-Nav system achieved the best performance, with the lowest navigation time (634 s) and distance (28 m), and the highest success rate (80%) and SPL (0.712). Removing $L_{\text{dict}}$ reduced performance due to the absence of generalized landmark positions, leading to frequent localization errors during reasoning. Excluding $L_{\text{obj}}^{\text{pred}}$ caused incorrect landmark detection, lowering SR to 40%. Removing the pruned topological map $\mathcal{M}'_{\text{tp}}$ further degraded results due to noisy position hypotheses in $I_{\text{SVAM}}$. The absence of the $EM$ caused the largest performance drop due to repeated hallucinations and failures to update state across steps. A common failure mode of HAM-Nav is incorrect localization, which leads to action loops or dead ends. This occurs when hand-drawn maps feature sparse landmark class labels (e.g., 'desk', 'plant') by the user, which reduces the semantic grounding for the VLM during position estimation. Replacing GPT-4o with Q2.5 family of models showed lower performance across all metrics. The Q2.5-72B model performed best among the open-source models but still underperformed relative to GPT-4o. This gap is due to limitations in Q2.5 models' reasoning stability and reduced grounding accuracy when aligning visual observations with textual prompts.

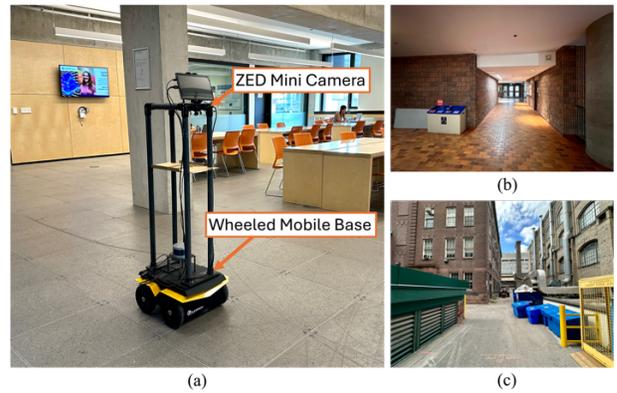

**Fig. 7.** (a) MH building (40 m × 43 m), (b) SF building (25 m × 15 m), (c) IA outdoor environment (35 m × 25 m).

### B. User Study in Real-World Environments

We conducted a user study to evaluate the feasibility and usability of HAM-Nav in real-world environments. We used the SR, SPL, NT and D as performance metrics. We also evaluated perceived usability of HAM-Nav against a VLN method using two standardized metrics: 1) the 5-point Likert System Usability Scale (SUS) [45], to provide an overall ease of use score, and 2) the Net Promoter Score (NPS) [46], to measure the likelihood of users recommending HAM-Nav based on their own experience. Before conducting the experiment, we received approval from the University of Toronto's (UofT) ethics committee (RIS Number: 47761).

*1) Real-World Environments:* Two structured indoor and one unstructured outdoor environment were used on the UofT campus, the: (1) Myhal Center for Engineering building (MH), Fig. 7(a), (2) Sandford Fleming building (SF), Fig. 7(b), and (3) Industrial Alley (IA), Fig. 7(c). The ground truth of MH is shown in Fig. 8(d).

*2) Mobile Robot:* A Jackal wheeled robot with a ZED Mini stereo camera was deployed. GMapping [47] was used to generate the occupancy grid during navigation for path planning. The A* algorithm and the TEB planner was used for global and local planning, respectively. Similar to the Ablation Study, we used GPT-4o as our VLM.

*3) Comparison Method:* MapGPT [27] is a VLN method that takes natural language navigation instructions as input. It serves as a baseline to evaluate performance differences between language-only instruction (MapGPT) and visual-language instructions using hand-drawn maps (HAM-Nav).

*4) Procedure:* Twenty participants (ages 22–42, μ: 30.2, σ: 5.7) were recruited, evenly split between STEM and non-STEM backgrounds to represent both technical and non-technical users. They were divided into two groups with equal distribution. One group evaluated HAM-Nav using hand-drawn maps, while the other evaluated MapGPT using verbal instructions. Each participant received a 5-minute tour of each environment to observe the spatial layout. Two start and goal positions were then randomly selected. HAM-Nav participants had 3 minutes to sketch the environment and robot path using an iPad and Apple Pencil. MapGPT participants were given a single attempt to verbally provide navigation instructions to simulate time-constrained scenarios. Each robot executed two trials per environment. After all trials, participants completed the SUS questionnaire from 1 (strongly disagree) to 5 (strongly agree) and answered the NPS question from 1 (not at all likely), to 10 (extremely likely) to recommend HAM-Nav/MapGPT.

TABLE III: Performance Metrics and Corresponding Results

| | | **MapGPT [27]** | | **HAM-Nav (ours)** | |
|---|---|---|---|---|---|
| **SUS Questionnaire** | | Median ($\tilde{x}$) | IQR | Median ($\tilde{x}$) | IQR |
| **S1** | I think that I would like to use [*MapGPT / HAM-Nav*] frequently to provide navigation instructions to a mobile robot. | 2 | 0 | 4 | 2 |
| **S2\*** | I found [*verbal commands / hand-drawn maps*] too complex for providing navigation instructions to a mobile robot. | 4 | 1 | 2 | 0 |
| **S3** | I thought [*MapGPT / HAM-Nav*] was easy to use. | 1 | 1 | 4 | 0 |
| **S4\*** | I believe I would need help from a technical person to use [*MapGPT / HAM-Nav*] effectively. | 2 | 1 | 1 | 1 |
| **S5** | I thought the time provided for [*giving verbal commands / drawing the hand-drawn map*] was sufficient for me. | 5 | 0 | 4 | 1 |
| **S6\*** | I thought the performance of [*MapGPT / HAM-Nav*] was inconsistent and did not meet my expectations based on my [*verbal command / hand-drawn map*]. | 3 | 2 | 2 | 1 |
| **S7** | I would imagine that most people would learn to use [*MapGPT / HAM-Nav*] very quickly. | 3 | 0 | 5 | 1 |
| **S8\*** | I found [*MapGPT / HAM-Nav*] to be very cumbersome to use. | 3 | 1 | 2 | 1 |
| **S9** | I felt confident using my memory to [*give verbal command / draw the map*] for navigation instructions. | 1 | 1 | 4 | 1 |
| **S10\*** | I needed to learn many things before I could start using [*MapGPT / HAM-Nav*]. | 1 | 0 | 1 | 1 |
| *\* Statements are negatively worded.* | **Avg. SUS Score** | 47.25 | | 79.5 | |
| **NPS Question:** How likely is it that you would recommend [*MapGPT / HAM-Nav*] to a friend or colleague? | **Ovr. NPS Score** | -80 | | 10 | |

| NT (s) ↓ | | D (m) ↓ | | SR ↑ | | SPL ↑ | |
|---|---|---|---|---|---|---|---|
| MapGPT | HAM-Nav | MapGPT | HAM-Nav | MapGPT | HAM-Nav | MapGPT | HAM-Nav |
| 1332 | 1018 | 59 | 45 | 64% | 77% | 0.523 | 0.703 |

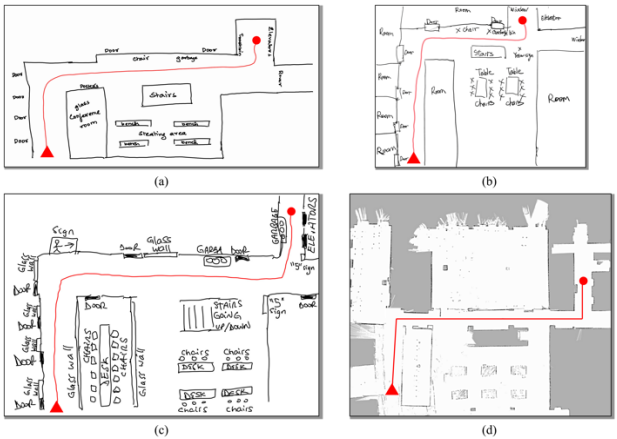

**Fig. 8.** Examples of hand-drawn maps in MH with low (a), medium (b), and high (c) landmark densities. (d) ground truth of MH. The starting and desired positions are denoted by red circles and triangles.

*4) Results:* Table III summarizes the SUS, NPS, NT, D, SR, and SPL results. HAM-Nav achieved a higher average SUS score (79.5) than MapGPT (47.25), corresponding to "Good–Excellent" and "Poor–OK" ratings, respectively [45]. MapGPT's lower SUS score was due to the cognitive load of generating complete verbal instructions in a single attempt. In contrast, HAM-Nav allowed gradual map creation with visual feedback, reducing memory burden. HAM-Nav participants reported high usability: frequent use (**S1**, $\tilde{x} = 4$, $IQR = 2$), low complexity (**S2**, $\tilde{x} = 2$, $IQR = 0$), ease of use (**S3**, $\tilde{x} = 4$, $IQR = 0$), and low cumbersomeness (**S8**, $\tilde{x} = 2$, $IQR = 1$). Most felt confident in using memory for sketching (**S9**, $\tilde{x} = 4$, $IQR = 1$), believed the provided time was sufficient (**S5**, $\tilde{x} = 4$, $IQR = 1$), and found the system learnable (**S7**, $\tilde{x} = 5$, $IQR = 1$). One participant expressed needing technical help (**S4**, $\tilde{x} = 1$, $IQR = 1$). HAM-Nav achieved an NPS of +10, while MapGPT scored –80, indicating strong user preference for HAM-Nav [46].

In real-world trials, HAM-Nav outperformed MapGPT in all metrics: NT (1018 s vs. 1332 s), D (45 m vs. 59 m), SR (77% vs. 64%), and SPL (0.703 vs. 0.523). HAM-Nav also generalized well across diverse sketch styles and landmark densities (Fig. 8). In contrast, MapGPT's requirement for single delivery of verbal instructions led to frequent omissions, mid-sentence corrections, and ambiguities, resulting in misinterpretation by the VLM and degraded navigation performance. A demonstration video of HAM-Nav is available at https://youtu.be/QGBRmpghMOM.

## VI. CONCLUSION

In this paper, we introduced the HAM-Nav architecture for mobile robot navigation using hand-drawn maps. Our approach uniquely leverages pre-trained VLMs for navigation. The novelty of HAM-Nav is in its robustness across varying environments and its ability to interpret diverse drawing styles without requiring the hand-drawn maps to be metrically accurate. The performance of HAM-NAV was validated through an ablation study as well as a user study. Results demonstrated that HAM-Nav can effectively navigate in both indoor and outdoor, single and multi-floor settings, with realistic landmarks. Current limitations include long navigation time due to the stop-and-go behavior of the robot as it waits for VLM outputs. Internet access is also required during operation to access VLM API. Future work will focus on running local VLMs for faster inference and extending HAM-Nav to support multi-robot systems.